\documentclass[sigconf, 9pt]{acmart}

\usepackage{hhline}
\usepackage{colortbl}
\usepackage{centernot}
\usepackage{pbox}
\usepackage{amsmath}
\usepackage{amsfonts}
\usepackage{url}
\usepackage{bm}
\usepackage{comment}
\usepackage{graphicx}
\usepackage{subfigure}
\usepackage{epstopdf}
\usepackage{multirow}
\usepackage{color}
\usepackage{tabu}
\usepackage{mdframed}
\usepackage{syntax}
\usepackage{fancyvrb}
\usepackage{makecell}
\usepackage{stfloats}

\usepackage[noend]{algpseudocode}
\usepackage{algorithmicx}
\usepackage{algpseudocode}
\usepackage{paralist}
\usepackage{stmaryrd}
\usepackage{tikz}
\usepackage[referable]{threeparttablex}
\usepackage{tabularx}
\usepackage{booktabs}
\usepackage{array}
\usepackage{fancyhdr}
\usepackage{float}
\usepackage{xspace}
\usepackage{pifont}
\usepackage{balance}

\usepackage{tablefootnote}
\usepackage[normalem]{ulem}

\usepackage{soul}
\usepackage{color}

\usepackage{tikz}

\usepackage{graphicx}
\usepackage{adjustbox}

\usepackage{float}

\newfloat{figtab}{htb}{fgtb}
\makeatletter
  \newcommand\figcaption{\def\@captype{figure}\caption}
  \newcommand\tabcaption{\def\@captype{table}\caption}
\makeatother

\definecolor{princetonorange}{RGB}{255,143,0}

\definecolor{lightgreen}{RGB}{198, 224, 183}

\definecolor{lightred}{RGB}{240, 205, 176}

\usepackage{geometry}
 \newcommand{\jing}[1]{\textcolor{newblue}{#1}}

 \definecolor{softblue}{rgb}{0.18,0.46,0.78}    
\definecolor{softgreen}{rgb}{0.03,0.60,0.27}

\usepackage[ruled,vlined,linesnumbered]{algorithm2e}

\definecolor{newblue}{RGB}{66, 173, 245}
\definecolor{cyan}{RGB}{213,229,255}
\definecolor{yellow}{RGB}{253, 243, 208}
\definecolor{orangeShallow}{RGB}{255,190,0}
\definecolor{ShallowYellow}{RGB}{249,241,204}
\definecolor{ShallowGreen}{RGB}{222,237,214}
\definecolor{ShallowOrange}{RGB}{250,229,212}
\definecolor{NormalGreen}{RGB}{51, 105, 30}
\definecolor{ShallowPurple}{RGB}{232, 223, 243}
\definecolor{redLight}{RGB}{255,128,114}
\definecolor{softRed}{RGB}{244, 113, 116}
\definecolor{newblue}{RGB}{66, 173, 245}
\definecolor{softpurple}{RGB}{249, 213, 246}

\definecolor{overfittingorange}{RGB}{255, 140, 37}

\usepackage{array}

\copyrightyear{2026}
\acmYear{2026}
\setcopyright{cc}
\setcctype{by}
\acmConference[DAC '26]{63rd ACM/IEEE Design Automation Conference}{July 26--29, 2026}{Long Beach, CA, USA}
\acmBooktitle{63rd ACM/IEEE Design Automation Conference (DAC '26), July 26--29, 2026, Long Beach, CA, USA}
\acmDOI{10.1145/3770743.3804053}
\acmISBN{979-8-4007-2254-7/2026/07}

\begin{document}

\title{RTL-BenchMT: Dynamic Maintenance of RTL Generation Benchmark Through Agent-Assisted Analysis and Revision}



\author{Jing Wang\textsuperscript{\dag}, Shang Liu\textsuperscript{\dag}, Hangan Zhou, Zhiyao Xie\textsuperscript{*}}
\affiliation{%
  \institution{Hong Kong University of Science and Technology}
  \city{Clear Water Bay}
  \country{Hong Kong}
}
\affiliation{%
  \institution{\textnormal{\texttt{\{jwangjw,sliudx,hzhoubu\}@connect.ust.hk, eezhiyao@ust.hk}}}
  \country{}
}

\pagestyle{plain}

\begin{abstract}

This paper introduces RTL-BenchMT, an agentic framework for dynamically maintaining RTL generation benchmarks.
Large Language Models (LLMs) assisted automated RTL generation is one of the most important directions in EDA research. However, current RTL benchmarks face two critical challenges: \textbf{(1) flawed cases in the benchmarks} and \textbf{(2) overfitting to the benchmarks}.
Both challenges are difficult to resolve purely by manual engineering effort.
To address these issues and \emph{systematically reduce human maintenance cost}, we propose an automated agentic framework, \textbf{RTL-BenchMT}.
RTL-BenchMT focuses on two key applications: \textbf{(1) automatically identifying and revising flawed benchmark cases} and \textbf{(2) automatically detecting and updating overfitting cases}.
With the assistance of RTL-BenchMT, we conduct a thorough, in-depth analysis of flawed and overfitting cases and produce a refined benchmark suite that will be open-sourced to the community.\looseness=-1

\end{abstract}




\maketitle

\let\oldthefootnote\thefootnote
\newcounter{tempcnt}
\setcounter{tempcnt}{\value{footnote}}

\renewcommand{\thefootnote}{\fnsymbol{footnote}}
\setcounter{footnote}{1}
\footnotetext{Corresponding author.} 

\stepcounter{footnote}
\footnotetext{Equal contribution.} 

\renewcommand{\thefootnote}{\oldthefootnote}
\setcounter{footnote}{\value{tempcnt}}

\section{Introduction}\label{sec:intro}

%

The EDA field is undergoing a transformative shift with the emergence of large language models (LLMs).
One of the most important applications is LLM-based automated RTL generation~\cite{lu2024rtllm, liu2024openllm, liu2023verilogeval, pinckney2024revisiting, pinckney2025comprehensive,liu2024rtlcoder, zhao2024mage, ho2024verilogcoder, cui2024origen, yao2024rtlrewriter, zhao2024codev,akyash2025decortl, ma2024verilogreader, pei2024betterv}, in which LLMs generate the desired RTL design from a natural-language description.
These works rely on open RTL benchmarks~\cite{lu2024rtllm, liu2023verilogeval, pinckney2025comprehensive} to measure functional correctness and compare models.
Despite tremendous advances, we find that existing RTL generation benchmarks suffer from two fundamental issues.


\textbf{Challenge 1. Flawed cases in the benchmark.}
{Existing RTL benchmarks inevitably contain flawed cases, which can misrepresent the true capability of LLMs.}
Some tasks exhibit inconsistencies between the design description and the reference testbench. Some others omit critical implementation details.
Such flaws can cause otherwise correct designs to be marked as failures, leading to unfair or misleading evaluation.
However, systematically identifying and revising flawed cases across large benchmarks is highly labor-intensive and requires substantial RTL and verification expertise.\looseness=-1


\textbf{Challenge 2. Overfitting on the benchmark.}
{Public RTL benchmarks are easily overfitted by LLMs, leading to increasingly over-optimistic performance results.}
Since benchmarks must be publicly available for research purposes, new LLM solutions tend to overfit the benchmark data~\cite{cohen2025forget} either by memorizing training examples or exploiting superficial patterns in the descriptions.
Detecting such overfitting is essential for a fair and unbiased evaluation of LLMs.
To the best of our knowledge, there is no practical framework that can automatically detect and quantify overfitting on RTL generation benchmarks.

\begin{figure}[t]
    \centering
    \includegraphics[width=0.95\linewidth]{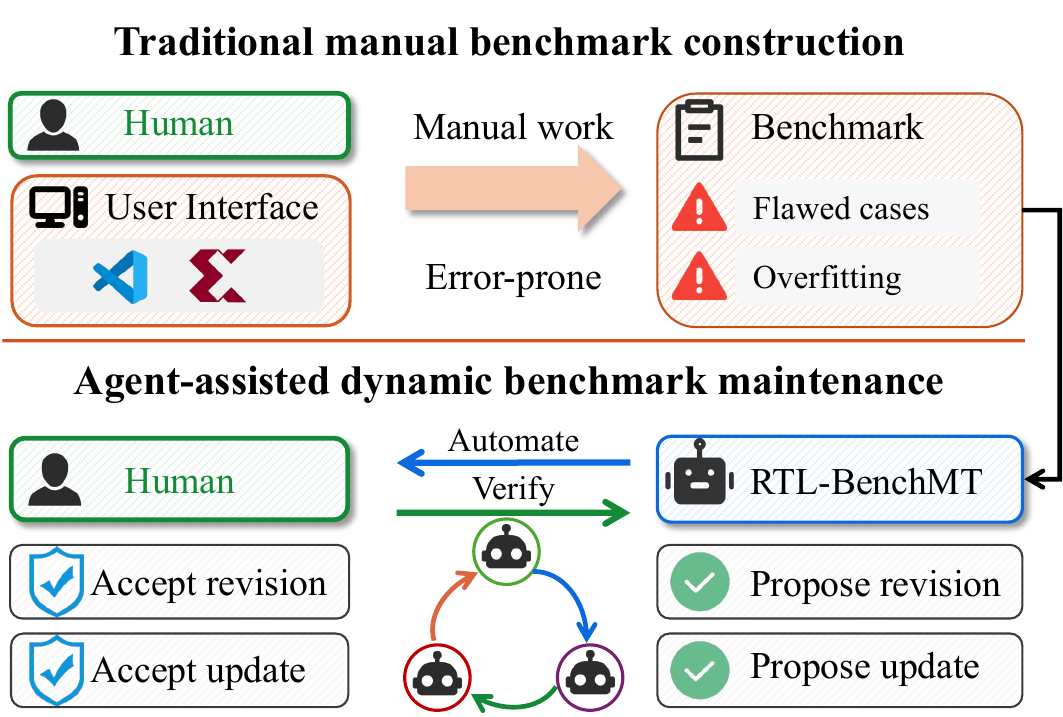}
    \vspace{-.1in}
    \caption{\emph{(1) Flawed cases} and \emph{(2) overfitting} are two significant challenges for RTL generation benchmarks. \emph{RTL-BenchMT} resolves the challenges by dynamically maintaining benchmarks. \emph{RTL-BenchMT} contributes in two important aspects: \uline{(1) automatically identifying and revising flawed cases} and \uline{(2) automatically detecting and updating overfitting cases.}}
    \label{fig:teaser-agentic-framework}
    \vspace{-.1in}
\end{figure}

\begin{figure*}[t!]
    \centering
    
    \includegraphics[width=0.99\linewidth]{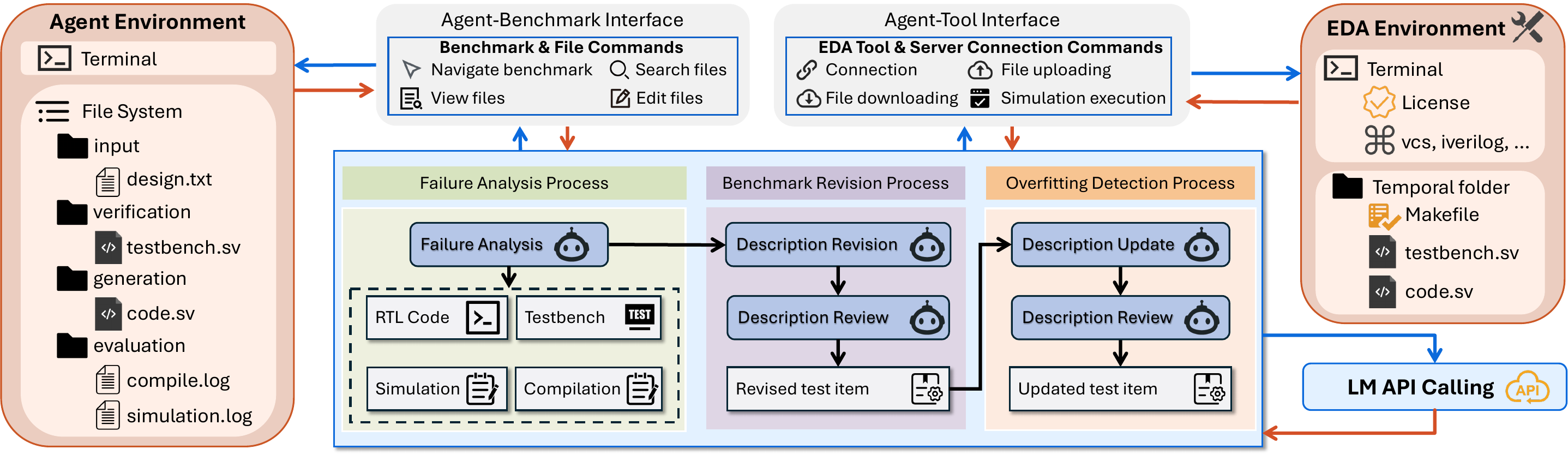}
    \vspace{-.1in}
    \caption{Overview of RTL-BenchMT agentic framework. The multi-agent system interacts with the environment through specified interfaces. The agent has three important automated phases: \emph{(1) failure analysis process, (2) benchmark revision process, and (3) overfitting detection.}}
    \label{fig:agentic-benchmark-analysis}
    \vspace{-.1in}
\end{figure*}

{In this work, we propose \textbf{RTL-BenchMT}, an agentic framework for dynamic maintenance of RTL generation benchmarks.}
As illustrated in Figure~\ref{fig:teaser-agentic-framework}, RTL-BenchMT organizes multiple specialized agents into an automated workflow that continuously analyzes benchmarks, revises flawed cases, and probes overfitting.
The framework supports two main capabilities: \textbf{(1) automatically identifying and revising flawed cases} and \textbf{(2) automatically detecting and updating overfitting cases}.
Human engineers shall remain in the loop to review RTL-BenchMT suggestions and approve the final benchmark updates.

To support effective maintenance, we design multiple specialized agents with distinct roles.
RTL-BenchMT orchestrates these agents through three processes.
\emph{Process 1} (\emph{failure analysis}) involves the \emph{failure analysis agent}, which inspects logs of failed cases and identifies potentially flawed benchmark cases.
\emph{Process 2} (\emph{benchmark revision}) revises the design descriptions of the identified cases.
In this process, a \emph{description revision agent} first proposes a candidate revision, and a \emph{description review agent} validates the revision according to strict rules.
\emph{Process 3} (\emph{overfitting detection}) tackles overfitting by rewriting design descriptions without changing their semantics; this process involves a \emph{description update agent} and a \emph{description review agent}.
Each agent is implemented under an \emph{iterative reasoning} paradigm, where the agent follows a three-step loop: \emph{generating thought}, \emph{taking action}, and \emph{obtaining observation}.

\textbf{Flawed case identification and revision}. RTL-BenchMT automatically pinpoints problematic tasks and proposes refined descriptions.
With the help of RTL-BenchMT, we propose a set of \textbf{refined benchmarks}, which will be open-sourced to the public.
Using \emph{Process 1} and \emph{Process 2}, RTL-BenchMT runs multiple LLMs on the benchmarks and aggregates consistently failing cases. Then RTL-BenchMT drives the revision agent to identify the flawed cases. Then, the analysis agent will propose a revision of the flawed cases. Finally, the review agent will validate the revision with strict rules.\footnote{We open-source the revised cases in the benchmarks in \url{https://github.com/hkust-zhiyao/RTL-BenchMT.git}.}

\textbf{Overfitting detection and updating.} RTL-BenchMT rewrites descriptions to expose possibly overfitting models that rely on superficial patterns.
\emph{Process 3 (overfitting detection process)} controls description-update and review agents that generate semantically equivalent descriptions. The LLMs will be evaluated and compared based on updated descriptions. A model is considered overfitted if it passes on the original description but fails on a rewritten one. This simple criterion provides an automatic per-case and per-model signal of overfitting strength, while the rewritten descriptions also increase benchmark diversity for future evaluations.

The rest of this paper is organized as follows.
Sec~\ref{sec:rtl-benchmt-framework} introduces the RTL-BenchMT framework and agent design.
Sec~\ref{sec:analysis-prompt-ambiguity} discusses identified flawed cases and the corresponding revision strategies.
Sec~\ref{sec:quantative-results} provides quantitative results on flawed-case identification and overfitting detection.

\section{RTL-BenchMT Agentic Framework}
\label{sec:rtl-benchmt-framework}

In this section, we introduce \textbf{RTL-BenchMT}, an agentic framework for dynamic maintenance of RTL generation benchmarks, as shown in Figure~\ref{fig:agentic-benchmark-analysis}.
In the following section, we first provide an overview of \emph{RTL-BenchMT} (Section~\ref{sec:overview}), including both the execution processes and agents. Then we provide a detailed introduction of techniques utilized in \emph{application 1: flaw identification and revision} (Section~\ref{sec:flawed-identification}), and \emph{application 2: overfitting detection and updating} (Section~\ref{sec:overfit-detection}). Finally, we introduce the infrastructures (Section~\ref{sec:details-rtlrev}), including the environments and interfaces.

\subsection{Overview of RTL-BenchMT} \label{sec:overview}

The RTL‑BenchMT framework consists of three main processes, as illustrated in Figure~\ref{fig:agentic-benchmark-analysis}. We first provide an overview of the three main processes: \emph{(1) failure analysis process}, \emph{(2) benchmark revision process}, and \emph{(3) overfitting detection process}. 
Within the framework, the \emph{manager agent} orchestrates the three core processes to support the two key applications.
Specifically, \emph{(1) The failure analysis process} is in charge of identifying flawed cases in benchmarks. In this process, \texttt{failure analysis agent} performs the core task: \emph{identifying the flawed cases} by our curated \texttt{analysis reasoning template}.
\emph{(2) The benchmark revision process} revises the flawed cases based on the identification results. In this process, \texttt{description revision agent} will first propose a revision, and \texttt{description review agent} will validate the revised description with strict \texttt{semantic rules}.
\emph{(3) The overfitting detection process} detects overfitting cases through \texttt{description updating strategies}. In this process, the \texttt{description update agent} first generates description variations by modifying the format while preserving the original semantics and functionality. The LLMs' performance on the description variation will reveal instances of overfitting.

\begin{figure*}
    \vspace{-.4in}
    \centering
    \includegraphics[width=0.99\linewidth]{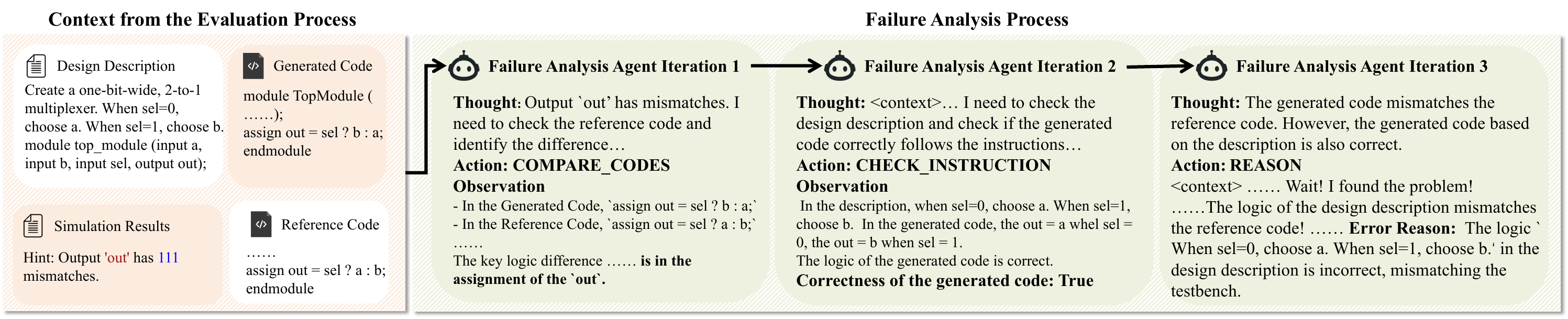}
    \vspace{-.1in}
    \caption{Example of automated flawed cases identification.}
    \label{fig:example-agentic-framework}
    \vspace{-.1in}
\end{figure*}

\textbf{Agents of RTL-BenchMT.}
All agents in the \emph{RTLBench-MT} framework follow a basic \emph{iterative reasoning} paradigm. At each iteration $i$, the agent performs the three-step actions: \texttt{(1) generating thought, (2) taking action}, and \texttt{(3) obtaining observation}.
Each agent has an \texttt{Action} list: `COMPARE\_CODES,'  `REASON,' `CHECK\_INSTRUCTION,' etc. At each iteration $i$, the agent chooses an action (denoted as $\mathcal{A}_i$) based on the thought (denoted as $\mathcal{T}_i$), then receives an observation (denoted as $\mathcal{O}_i$) from the environment (\texttt{Env}). Given the observation from the previous step, $\mathcal{O}_{i-1}$, the agent generates the thought $\mathcal{T}_i$ for the next step. Then the agent takes the action $\mathcal{A}_i$ for the next step based on the thought $\mathcal{T}_i$. After the action is taken, the new observation $\mathcal{O}_i$ will be returned. Such a process can be formulated as follows:

\setlength{\abovedisplayskip}{4pt}
\setlength{\belowdisplayskip}{4pt}
\begin{align}
    \mathcal{T}_{i} = \texttt{Agent} (\mathcal{O}_{i-1});~
    \mathcal{A}_i = 
    \texttt{Agent} (\mathcal{T}_i); ~ \mathcal{O}_{i} = \texttt{Env} (\mathcal{A}_{i})
\end{align}

Usually, the initial observation $\mathcal{O}_0$ is the evaluation results.
Take the \emph{(1) failure analysis process} as an example, the initial observation $\mathcal{O}_0$ for \texttt{analysis agent} is the evaluation results of the generated RTL code from LLMs. 
The \emph{RTL-BenchMT} workflow will first execute a standardized evaluation across a diverse suite of LLMs and collect the most common failure cases. Such failure cases will be propagated as observation $\mathcal{O}_0$ to the \texttt{failure analysis agent}. The agent will first generate thoughts $\mathcal{T}_1$ based on the evaluation results and decide the next actions $\mathcal{A}_1$ (e.g., `CHECK\_INSTRUCTION,' which reads the file of design description) and returns detailed contents of the design description $\mathcal{D}_\text{design}$ as a new observation $\mathcal{O}_1$.

\subsection{Application 1: Flaw Identification and Revision} \label{sec:flawed-identification}

The automated flawed case identification relies on effective failure case analysis, which is challenging and time-consuming.
The RTL-BenchMT framework addresses the significant challenge by automating the failure analysis process.
Figure~\ref{fig:agentic-benchmark-analysis} demonstrates the automated process, where the \emph{failure analysis process} and \emph{benchmark revision process} are executing the task.
The \textbf{RTL-BenchMT} workflow will first execute a standardized evaluation across a diverse suite of LLMs and collect the most common failure cases. Such failure cases will be propagated to the failure analysis agent.

The \textbf{failure analysis agent} follows a three-step \emph{thought-action-observation} reasoning paradigm.
The analysis agent applies iterative observation and reasoning on the failure cases to identify the flaw within the design description.
We curate a set of actions and an iterative reasoning paradigm for the analysis agent.

\vspace{-.1in}
\begin{algorithm}[ht]\small
    \caption{Analysis Reasoning Template}
    \label{alg:three-step-reason-template}
    \textbf{Iteration 1.} Code Mismatch Analysis\;
    $\mathcal{O}_0$ = \texttt{Simulation results}; \texttt{Action} $\mathcal{A}_1$ = \texttt{COMPARE\_CODES}\;
    \texttt{Observation} $\mathcal{O}_1$ = \{\texttt{Generated Code}, \texttt{Reference Code}\}\;
    \textbf{Iteration 2.} Code Correctness Analysis\;
    \texttt{Thought} $\mathcal{T}_2$ = \texttt{Agent} ($\mathcal{O}_1$);
    \texttt{Action} $\mathcal{A}_2$ = \texttt{CHECK_INSTRUCTION}\;
    \texttt{Observation} $\mathcal{O}_2$ =  Design Description $\mathcal{D}_\text{design}$\; 
    \textbf{Iteration 3.} Description mismatch detection\;
    \texttt{Thought} $\mathcal{T}_3$ = Agent($\mathcal{O}_2$)\;
    \If{\text{Code is correct} in $\mathcal{T}_3$}{
        \texttt{Action} $\mathcal{A}_3$ = \texttt{REASON}; \texttt{Context}= \{$\mathcal{O}_1, \mathcal{O}_2, \mathcal{O}_3$\} \; \texttt{Observation} $\mathcal{O}_3$ = Agent(\texttt{Context}) = \{\texttt{IS\_FLAW, $\mathcal{R}_\text{flaw}$}\}\;
    }
    
\end{algorithm}
\vspace{-.1in}

\textbf{Analysis reasoning template.} Algorithm~\ref{alg:three-step-reason-template} illustrates the three-iteration template. In the first iteration (\emph{iteration 1}), the agent analyzes the mismatches between the codes (observation $O_1$). Then, the agent checks the design description (observation $O_2$ in \emph{iteration 2}) 
to verify if the generated code correctly implements the design specified in the description (if `code is correct' in $T_3$). If the generated code is identified as correct, then the defects of LLMs are eliminated from the failure reason. Finally, the agent analyzes (Action $A_3$ = \texttt{REASON}) based on the previous observations (\texttt{Context}=\{$O_1, O_2, O_3$\} in \emph{iteration 3}). In this iteration, the agent will specifically focus on the design description and testbench to identify the mismatch. We provide an example in the following part to demonstrate the detailed results.

\textbf{Example.} Figure~\ref{fig:example-agentic-framework} illustrates the automated failure analysis process.
At the input, the design description contains a mismatch in logic from the reference code in the testbench.
At iteration 1, the failure analysis agent compares the generated code and the reference code. The agent identifies the mismatch between the generated code and the reference code.
At iteration 2, then, the agent decides to thoroughly analyze the generated code based on the original description generation.
In this iteration, the agent concludes that the generated code correctly implements the design based on the design description. Thus, the agent excludes the possibility of LLMs' ability defect.
At iteration 3, this observation finally motivates the analysis agent to compare the design description and the reference code. Finally, the agent identifies the mismatching logic between the design description and the testbench.

Based on the key observation from the failure analysis, the \textbf{Benchmark Revision Process} updates the design description accordingly. The revision agent and the review agent collaborate to propose a revision. 
In this process, the \emph{review agent} checks the revision with two strict rules. (1) The revision should not destroy the basic semantics. Under this requirement, the revision can slightly change the functionality to align with the testbench's requirements.
(2) The revision should not directly leak the information from the testbench. For example, we have observed that the revision may directly leak code samples from the reference code as a shortcut for LLMs under test. This can also be viewed as a special case of revision destroying the semantics.
Finally, the engineer can further validate the revision.

\subsection{Application 2: Overfitting Detection and Updating} \label{sec:overfit-detection}

 To detect the overfitting cases, we design \emph{Process 3.} \emph{overfitting detection}. In this process, \emph{description update} agent creates more variance of design descriptions by rewriting the original description while keeping its semantics and functionality. A practical option is to shift the description styles.
 We pre-define a set of style templates for the \emph{update agent} as a reference, e.g., `Technical/Formal Style,' `Educational/Tutorial Style,' `Problem/Task Solving Style,' `Specification/Documentation Style,' etc.
Define the original description as \texttt{Desp}, the updated description as $\texttt{Desp}'$. The update process can be formulated as follows:
\begin{align}
    \mathcal{T}_\text{Plan} = \texttt{Agent}(\mathcal{D}_\text{design}, \{\texttt{Style}\});~ \mathcal{D}_\text{variation} = \texttt{Agent}(\mathcal{T}_\text{Plan})
\end{align}
The \emph{description update agent} will be provided a set of styles, denoted as $\{S_1, S_2, ...\}$. Then the agent will generate the $\texttt{Plan}$ for updating. Finally, the \emph{update agent} will update the design description based on the $\texttt{Plan}$.

\subsection{Details of RTL-BenchMT}
\label{sec:details-rtlrev}
This subsection describes the \textbf{runtime environments} and \textbf{tool interfaces} that RTL-BenchMT exposes to LLM agents.
From the agent's perspective, we distinguish between the environment where the agent itself runs and the environment where commercial EDA tools are executed.

\textbf{Environments.}
The LLM agents and orchestration scripts run in a lightweight host runtime, which we refer to as the \emph{agent environment}.
Commercial EDA tools are installed inside a Docker image running on a license server; we refer to this isolated runtime as the \emph{EDA environment}.
The agent never invokes EDA binaries directly.
Instead, it issues high-level tool calls that internally start or attach to a Docker container, execute compilation and simulation commands inside the container, and collect log files or waveforms as outputs.

\textbf{Interfaces.}
RTL-BenchMT exposes two groups of tools to the agent: \emph{benchmark-related} tools and \emph{EDA-related} tools.
Benchmark-related tools form the \textbf{agent--benchmark interface}, which allows the agent to locate a test case, search for files, view their contents, and edit RTL, testbenches, or textual descriptions.
EDA-related tools form the \textbf{agent--tool interface}, which mediates all communication with the Dockerized EDA environment.
Typical operations include establishing an EDA session, uploading the current design directory, invoking compilation and simulation commands in Docker, and retrieving the resulting reports or waveforms.

These interfaces constitute the atomic operations in the agent \texttt{Action} set.
Table~\ref{tab:action-interface} lists the most commonly used actions and their corresponding interface calls.
For example, when the agent chooses the \texttt{EVALUATION} action, it sequentially invokes five interfaces: (1) connect to the EDA environment, (2) upload the design directory, (3) run compilation, (4) run simulation, and (5) download the simulation results for further analysis.

\begin{table}[t]
    \vspace{-.4in}
    \centering
    \resizebox{.99\linewidth}{!}
    {\begin{tabular}{c|c}
    \hline
        \cellcolor{softRed!14}{\texttt{Action}} & \cellcolor{softgreen!14}{Interface usage \& combination} \\
        \hline
        \hline
         \multirow{2}{*}{\texttt{COMPARE_CODES}} & \texttt{1. view_file(generated\_code.sv)}, \\
         & 2. \texttt{view_file(testbench.sv)} \\ 
         \hline
    \texttt{CHECK_INSTRUCTION} & \texttt{1. view_file(description.txt)} \\
    \hline
    \multirow{3}{*}{\texttt{EVALUATION}} & \texttt{1. connect(), 2. upload(design\_folder)}\\
    & \texttt{3. compile(), 4. simulate()}\\
     & \texttt{5. download(results)} \\
         \hline
         \texttt{REVISE} & \texttt{1. edit_file(revised\_description,} 
         \\ 
         & \texttt{description.txt)}\\
         \hline
    \end{tabular}}
    \caption{Illustration of commonly used agent \texttt{Action}s and their interface calls.}
    \label{tab:action-interface}
    \vspace{-.3in}
\end{table}

\section{Analysis on Identified Flawed Cases}
\label{sec:analysis-prompt-ambiguity}

In this section, we provide a qualitative analysis of the identified flawed cases. For each example, we also provide suggested refinements of the cases.
The flaws are mainly caused by ambiguity of descriptions, categorized into three types:

\begin{itemize}
    \item \emph{Syntax Ambiguity.} Design descriptions may accidentally include contents that lead to LLMs' syntax errors. 
    \item \emph{Functional Behavior Ambiguity.} Design description may contain unclear descriptions of the functional behavior, especially for the detailed operations.
    \item \emph{Diagram Ambiguity}. The design description may contain diagrams. However, the manually written diagrams are prone to specifying incorrect or ambiguous logic.
\end{itemize}

\begin{figure}[t!]
    \vspace{-.4in}
    \centering
    \includegraphics[width=0.99\linewidth]{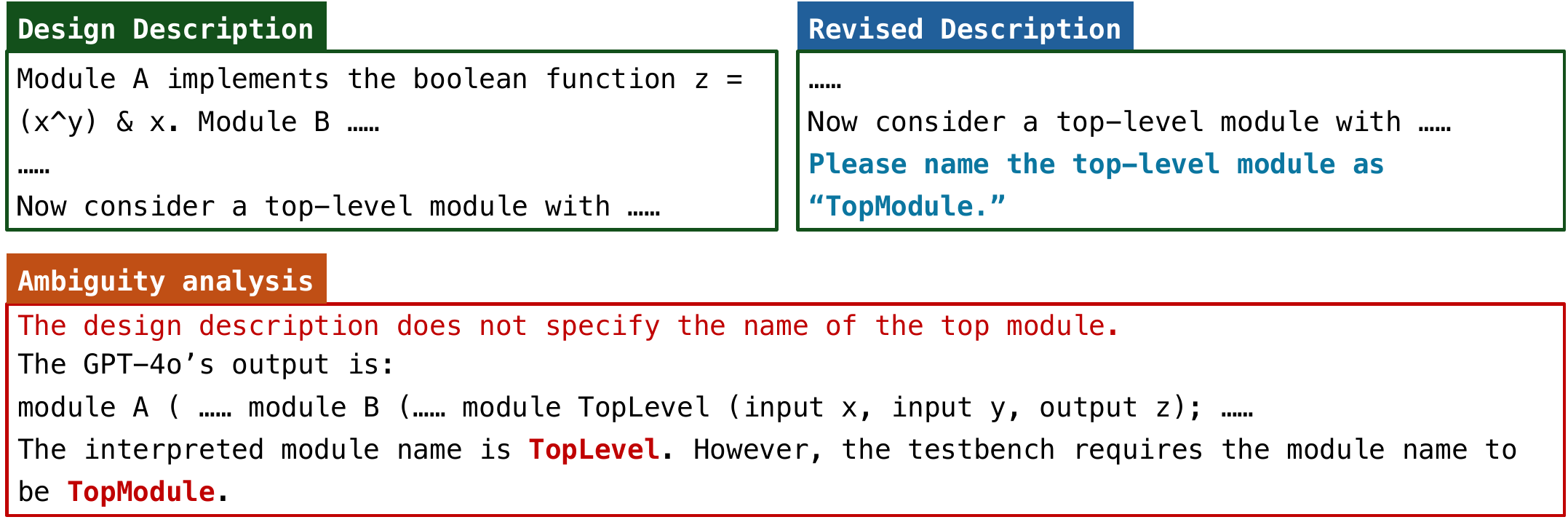}
    \vspace{-.1in}
    \caption{\emph{Undefined module name}. The testbench requires the module name to be `TopModule,' but this is not specified in the design description.}
    \label{fig:131-not-name}
    \vspace{-.2in}
\end{figure}

\begin{table*}[!b]
    \centering
    \caption{Functional accuracy of different models on different benchmarks.}
    \vspace{-.1in}
    \label{tab:Test-Benchmark}
    \resizebox{.99\textwidth}{!}
    {\begin{tabular}{c|c||c|c|c|c|c|c|c|c|c|c|c|c}
    \hline
    Type & Models &
         \multicolumn{2}{c|}{VerilogHuman v1} & \multicolumn{2}{c|}{VerilogHuman v2} & \multicolumn{2}{c|}{VerilogMachine v1} & \multicolumn{2}{c|}{RTLLM v1.1} & \multicolumn{2}{c|}{{cid02} in CVDP} & \multicolumn{2}{c}{{cid03} in CVDP} \\
         \hline
       & & Syntax & Function & Syntax & Function & Syntax & Function & Syntax & Function & Syntax & Function & Syntax & Function \\
         \hline
         \hline
         \multicolumn{14}{c}{Original pass@1 functional accuracy}\\
         \hline
         \hline
          \multirow{3}{*}{Commercial} & GPT-3.5 & 73.7\% & 32.7\% & 53.2\% & 26.3\% & 25.0\% & 16.0\% & 80\% & 50\%  & - & - & - & - \\
          \cline{3-14}

           & GPT-4o-mini & 89.1\% & 51.3\% & 96.2\% & 51.9\% & 87.8\% & 59.6\% & 88\% & 56\% & - & 10.6\% & - & 18.1\% \\
           \cmidrule{2-14}
          
           \multirow{2}{*}{model} & \cellcolor{softRed!20}{GPT-4o} & \cellcolor{softRed!20}{96.8\%}& \cellcolor{softRed!20}{67.3\%} & \cellcolor{softRed!20}{96.2\%} & \cellcolor{softRed!20}{68.8\%} & \cellcolor{softRed!20}{96.5\%} & \cellcolor{softRed!20}{74.1\%} & \cellcolor{softRed!20}{92\%} & \cellcolor{softRed!20}{68\%} & \cellcolor{softRed!20}{-} & \cellcolor{softRed!20}{19.1\%} & \cellcolor{softRed!20}{-} & \cellcolor{softRed!20}{35.1\%} \\

           \cline{3-14}

          & \cellcolor{softRed!20}{Claude-3.7} & \cellcolor{softRed!20}{98.1\%} & \cellcolor{softRed!20}{\textbf{80.1\%}} & \cellcolor{softRed!20}{98.1\%} & \cellcolor{softRed!20}{\textbf{76.3\%}} & \cellcolor{softRed!20}{92.3\%} & \cellcolor{softRed!20}{\textbf{76.9\%}} & \cellcolor{softRed!20}{90\%} & \cellcolor{softRed!20}{\textbf{70.0\%}} & \cellcolor{softRed!20}{-} & \cellcolor{softRed!20}{\textbf{21.3\%}} & \cellcolor{softRed!20}{-} & \cellcolor{softRed!20}{\textbf{39.0\%}} \\
          \midrule
          
          
          
            & \multirow{1}{*}{QwQ-32B} & \multirow{1}{*}{53.2\%} & \multirow{1}{*}{44.2\%} & \multirow{1}{*}{61.5\%} & \multirow{1}{*}{55.1\%} & \multirow{1}{*}{74.4\%} & \multirow{1}{*}{61.5\%} & \multirow{1}{*}{28\%} & \multirow{1}{*}{26\%} & \multirow{1}{*}{-} & \multirow{1}{*}{8.5\%} & \multirow{1}{*}{-} & \multirow{1}{*}{14.3\%} \\
          
          \cline{3-14}
          
          \multirow{-2}{*}{Open} & LlaMA-70B & 78.8\% & 47.4\% & 80.1\% & 42.9\% & 78.8\% & 47.4\% & 54\% & 34\% & - & 10.6\% & - & 12.3\% \\

          \cline{3-14}

          \multirow{-2}{*}{model} & LlaMA-405B & 75\% & 48.7\% & 91.7\% & 60.9\% & 81.4\% & 62.2\% & 67.3\% & 46.8\% & - & 16.0\% & - & 27.3\% \\
          \cline{3-14}
          
          \hline
          \hline
          \multicolumn{14}{c}{Functional accuracy after \textbf{RTL-BenchMT} fixes flawed cases in benchmark}\\
          \midrule
          
            \multirow{2}{*}{Commercial} & \cellcolor{softblue!14} & \cellcolor{softblue!14} & \cellcolor{softblue!14} 71.1\% & \cellcolor{softblue!14} & \cellcolor{softblue!14} 71.4\%  & \cellcolor{softblue!14} & \cellcolor{softblue!14} {77.6\%}  & \cellcolor{softblue!14}  & \cellcolor{softblue!14} \multirow{1}{*}{70\%} & \cellcolor{softblue!14} & \cellcolor{softblue!14} 20.2\%  & \cellcolor{softblue!14}  & \cellcolor{softblue!14} 33.8\%  \\
           & \cellcolor{softblue!14} \multirow{-2}{*}{GPT-4o} &  \cellcolor{softblue!14}\multirow{-2}{*}{98.7\%} & \cellcolor{softblue!14} (\textcolor{NormalGreen}{+3.8\%}) & \cellcolor{softblue!14} \multirow{-2}{*}{98.1\%} &  \cellcolor{softblue!14}(\textcolor{NormalGreen}{+2.6\%}) &\cellcolor{softblue!14} \multirow{-2}{*}{96.5\%}  &\cellcolor{softblue!14}  (\textcolor{NormalGreen}{+3.5\%}) & \cellcolor{softblue!14} \multirow{-2}{*}{92\%} & \cellcolor{softblue!14} (\textcolor{NormalGreen}{+2.0\%}) & \cellcolor{softblue!14}\multirow{-2}{*}{-}  & \cellcolor{softblue!14} (\textcolor{NormalGreen}{+1.1\%}) & \cellcolor{softblue!14} \multirow{-2}{*}{-}  & \cellcolor{softblue!14} (\textcolor{red}{-1.3\%}) \\
          
          \cmidrule{2-14}
           \multirow{2}{*}{model} & \cellcolor{softblue!20} & \cellcolor{softblue!20} & \cellcolor{softblue!20} \textbf{79.8\%}  & \cellcolor{softblue!20} & \cellcolor{softblue!20} \textbf{76.9\%}  & \cellcolor{softblue!20} & \cellcolor{softblue!20} \textbf{78.2\%}    & \cellcolor{softblue!20}  & \cellcolor{softblue!20} \textbf{70\%} & \cellcolor{softblue!20} &  \cellcolor{softblue!20} \textbf{22.4\%}  & \cellcolor{softblue!20} & \cellcolor{softblue!20} \textbf{39.0\%}   \\
          
            & \cellcolor{softblue!20} \multirow{-2}{*}{Claude-3.7} & \cellcolor{softblue!20} \multirow{-2}{*}{99.3\%} & \cellcolor{softblue!20} (\textcolor{red}{-0.3\%}) & \cellcolor{softblue!20} \multirow{-2}{*}{98.1\%}  & \cellcolor{softblue!20}(\textcolor{NormalGreen}{+0.6\%}) & \cellcolor{softblue!20} \multirow{-2}{*}{92.3\%} &  \cellcolor{softblue!20}(\textcolor{NormalGreen}{+1.3\%})   & \cellcolor{softblue!20} \multirow{-2}{*}{90\%}  & \cellcolor{softblue!20} (\textcolor{gray}{+0.0\%}) & \cellcolor{softblue!20} \multirow{-2}{*}{-} & \cellcolor{softblue!20} (\textcolor{NormalGreen}{+1.1\%}) &   \cellcolor{softblue!20} \multirow{-2}{*}{-} & \cellcolor{softblue!20} (\textcolor{gray}{+0.0\%})  \\

          \bottomrule
        
    \end{tabular}}
\end{table*}

\begin{figure}[b!]
    \vspace{-.1in}
    \centering
    \includegraphics[width=0.99\linewidth]{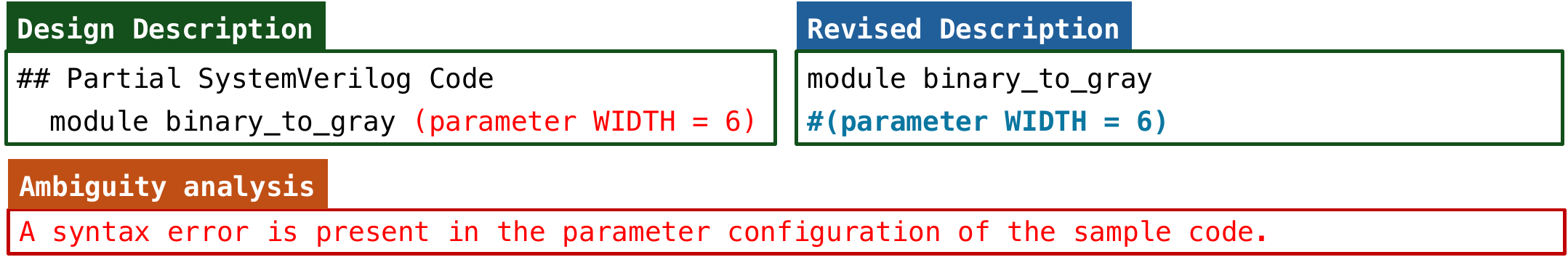}
    \vspace{-.1in}
    \caption{\emph{Code sample with syntax error}. Task ID: binary\_to\_gray\_0001 of \emph{cid002} in CVDP benchmark. The code sample in the design description contains a syntactically incorrect parameter configuration that misleads LLMs into generating the same syntax error.}
    \label{fig:cid002-binary-to-gray-0001-syntax-prompt}
\end{figure}

\subsection{Syntax ambiguity} 

We introduce three identified situations of syntax ambiguity: (1) \emph{undefined module name}, (2) \emph{unclear port type}, and (3) \emph{syntax errors in code example}.

\textbf{Undefined module name} refers to the situation where the design description only specifies the functional logic while ignoring the name of the top module, which is required in the testbench. 
For example, in VerilogEval Human v2~\cite{liu2023verilogeval}, all the test cases within the benchmark require generating the design with the name of `TopModule.' 
However, some specific test cases do not provide the module name in the design description.
Figure~\ref{fig:131-not-name} presents an example of an undefined module name.
The GPT-4o interprets the module name as `TopLevel,' with a correctly implemented logic and only a different name for the top module. To fix the ambiguity, we can mention in the description that the top-level module should be named as `TopModule' rather than `TopLevel.'

\textbf{Unclear port type} refers to the fact that the design description may not clearly specify the output port type (wire or register) of the module. Thus, the LLMs will fail the task if they interpret a different port type. For example, the LLM may try to assign the output port value in a sequential block because it interprets the output port as a register, which is required as a wire in the testbench.

\textbf{Syntax errors in the code example} are present in the description containing a code example as hints. However, the code example contains errors, and LLMs will usually directly copy most of it, including the syntax errors. Figure~\ref{fig:cid002-binary-to-gray-0001-syntax-prompt} illustrates the obvious syntax error present in the given template code example.
The code example specifies a parameter configuration, but with syntax errors. We can resolve this ambiguity by correcting the code sample's syntax errors.

\begin{figure}[t!]
    \centering
    \includegraphics[width=.99\linewidth]{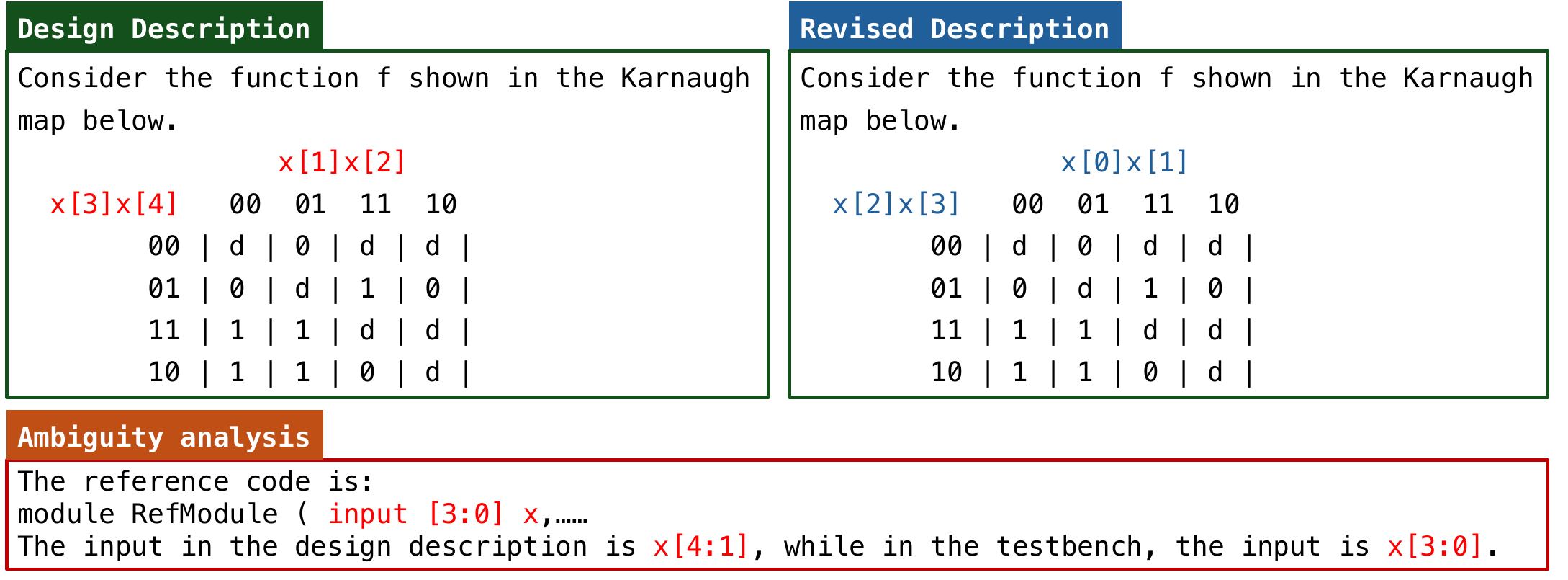}
    \vspace{-.1in}
    \caption{\emph{Diagram ambiguity}. Task ID: HumanEval v2, 116 m2014\_q3. The specified input is `x[4:1]', while in the reference code, the input is `x[3:0].'}
    \label{fig:116-input-mismatch}
    \vspace{-.2in}
\end{figure}

\subsection{Functional behavior ambiguity} 

Functional behavior ambiguity includes both sequential and combinational logic. Such behavior ambiguity is usually related to
the detailed operations, like bit-level operations or conditioned branches. We list three representative situations: (1) \emph{register initialization}, (2) \emph{trigger condition}, and (3) \emph{missing implementation}.

\textbf{Register initialization} ambiguity exists in some designs with the register output port or intermediate registers. For example, in the VerilogEval~\cite{liu2023verilogeval, pinckney2024revisiting} benchmark, some tasks require the output port of a design to be a register. However, the testbenches require the output to be initialized to fixed values (for e.g., 0), which is not specified in the design description. As a result, LLMs that generate correct logic can still fail due to missing initial assignments in the testbench.

\textbf{Trigger condition ambiguity} refers to the trigger condition of sequential blocks. Some design descriptions do not specify whether a sequential block should be clocked on the clock’s posedge with a synchronous reset (reset sampled on the clock) or triggered asynchronously on the reset’s posedge (asynchronous reset). LLMs typically produce correct internal logic but miss whether the `reset' edge should be added to the trigger condition. Usually, LLMs have to choose a trigger condition by default randomly. Thus, the generated designs always fail due to this missing condition specification.

\textbf{Missing implementation} refers to the missing implementation of smaller modules. For example, some design descriptions require LLMs to generate a larger module from predefined smaller modules, which LLMs often interpret as already implemented. However, such pre-defined smaller modules are not provided in the testbench. LLMs are required to actually implement the smaller modules first.

\begin{table*}[t!]
    \vspace{-.4in}
    \centering
    \resizebox{.99\textwidth}{!}
    {\begin{tabular}{c|c c | c c|c c|c c|c c|c c}
    \hline
        \multirow{2}{*}{Benchmarks} & \multicolumn{2}{c|}{VerilogEval} & \multicolumn{2}{c|}{VerilogEval} & \multicolumn{2}{c|}{VerilogEval} & \multicolumn{2}{c|}{RTLLM}  & \multicolumn{2}{c|}{CVDP} & \multicolumn{2}{c}{CVDP}  \\
         & \multicolumn{2}{c|}{Machine v1} & \multicolumn{2}{c|}{Human v1} & \multicolumn{2}{c|}{Human v2} & \multicolumn{2}{c|}{v1.1} & \multicolumn{2}{c|}{cid002} & \multicolumn{2}{c}{cid003} \\
         \hline
         & number & percentage & number & percentage & number & percentage & number & percentage & number & percentage & number & percentage \\
         \hline
         \multicolumn{13}{c}{Cases that LLMs constantly fail} \\
         
         \midrule Total & \cellcolor{softblue!40}{13/143} & \cellcolor{softblue!40}{9.1\%} & \cellcolor{softblue!50}{21/156} & \cellcolor{softblue!50}{13.5\%} & \cellcolor{softblue!50}{18/156} & \cellcolor{softblue!50}{11.5\%} & \cellcolor{softblue!30}{8/50} & \cellcolor{softblue!30}{16\%} & \cellcolor{softblue!20}{11/94} & \cellcolor{softblue!20}{11.7\%} & \cellcolor{softblue!50}{28/77} & \cellcolor{softblue!50}{36.4\%}  \\
         \bottomrule

         \multicolumn{13}{c}{Identified flawed cases}\\

   \midrule Total &  \cellcolor{softRed!40}{11/143} & \cellcolor{softRed!40}{7.7\%} & \cellcolor{softRed!60}{14/156} & \cellcolor{softRed!60}{9.0\%} & \cellcolor{softRed!40}{8/156} & \cellcolor{softRed!40}{5.1\%} & \cellcolor{softRed!30}{5/50} & \cellcolor{softRed!30}{10.0\%} & \cellcolor{softRed!30}{4/94} & \cellcolor{softRed!30}{4.3\%} & \cellcolor{softRed!30}{5/77} & \cellcolor{softRed!30}{6.5\%} \\

    \midrule Functional & \cellcolor{softRed!40}{7/143} & \cellcolor{softRed!40}{6.3\%} & \cellcolor{softRed!50}{10/156} & \cellcolor{softRed!50}{6.4\%} & \cellcolor{softRed!40}{5/156} & \cellcolor{softRed!40}{3.2\%} & \cellcolor{softRed!20}{3/50} & \cellcolor{softRed!20}{6.0\%} & \cellcolor{softRed!10}{2/94} & \cellcolor{softRed!10}{2.1\%} & \cellcolor{softRed!10}{1/77} & \cellcolor{softRed!10}{1.3\%} \\
    
    \midrule Syntax & \cellcolor{gray!10}{-/-} & \cellcolor{gray!10}{-} & \cellcolor{gray!10}{-/-} & \cellcolor{gray!10}{-} & \cellcolor{gray!10}{-/-} & \cellcolor{gray!10}{-} & \cellcolor{softRed!20}{2/50} & \cellcolor{softRed!20}{4.0\%} & \cellcolor{softRed!10}{1/94} & \cellcolor{softRed!10}{1.1\%} & \cellcolor{softRed!20}{3/77} & \cellcolor{softRed!20}{3.9\%} \\
    
    \midrule Diagram & \cellcolor{softRed!20}{4/143} & \cellcolor{softRed!20}{2.8\%} & \cellcolor{softRed!20}{4/156} & \cellcolor{softRed!20}{2.6\%} & \cellcolor{softRed!20}{3/156} & \cellcolor{softRed!20}{1.9\%} & \cellcolor{gray!10}{-/-} & \cellcolor{gray!10}{-} & \cellcolor{softRed!10}{1/94} & \cellcolor{softRed!10}{1.1\%} & \cellcolor{softRed!10}{1/77} & \cellcolor{softRed!10}{1.3\%} \\
    \bottomrule
    \end{tabular}}
    \caption{Statistics on benchmark ambiguity. The upper block reports, for each benchmark, the proportion of tasks that all evaluated LLMs consistently fail, and the lower block shows how many of these are manually confirmed as flawed and how they decompose into functional, syntax, and diagram ambiguities. Across benchmarks, roughly 5--10\% of tasks are ambiguous, and a large fraction of consistently failing cases are actually flawed.}
    \label{tab:benchmark-ambiguity-analysis}
    \vspace{-.3in}
\end{table*}

\subsection{Diagram ambiguity} 

Diagram ambiguity always exists in the design diagrams, which represent a classical specification format for RTL design.
However, the manually written diagrams are prone to specifying incorrect or ambiguous logic.
We present two representative diagrams for illustration: the KMap and FSM formats.

\textbf{KMap diagram ambiguity} refers to the unclear representation of a KMap.
There are two kinds of ambiguity problems: 1)  indication about the rows and columns (e.g., which input parts correspond to which columns or rows) is unclear or misleading, 2) the values do not match the testbench. For example, in Figure~\ref{fig:116-input-mismatch}, the input in the KMap should be x[0:3], but the testbench requires the input to be x[1:4]. We can resolve this ambiguity by aligning the input x with the testbench.

\textbf{FSM diagram ambiguity} usually exists in the transfer condition of states. For example, the manually written FSM usually misses a few transition conditions, leading to an incomplete FSM specification.
The effective way to fix the flaws is to identify the missing state conditions and add them back, following the styles of the original description.

\section{Quantative Analysis}\label{sec:quantative-results}


\subsection{Flawed Case Identification}\label{sec:flaw-case-identification}

In this section, we analyze the identified flawed cases across different benchmarks. Table~\ref{tab:benchmark-ambiguity-analysis} presents detailed statistics on flawed cases.
Human engineers carefully verify all the identified cases.
The table consists of two parts: 1) \emph{cases that LLMs constantly fail} and 2) \emph{identified flawed cases}. The number of part 1) obviously correlates with the number of identified flawed cases. Such an observation reveals that \emph{many failure cases of LLMs are due to the case flaw rather than the LLMs' defects}.

It's noticeable that each benchmark contains around 10\% of ambiguous test cases. Each benchmark reveals a different pattern of flaw distribution. For example, the flawed cases in VerilogEval~\cite{liu2023verilogeval} benchmarks are mainly related to functional behaviors and diagrams, with no syntax-related flaws. RTLLM~\cite{lu2024rtllm} flawed cases are not related to diagrams. \emph{cid002} and \emph{cid003} tasks in the CVDP~\cite{pinckney2025comprehensive} benchmark contain new flawed cases related to the syntax.
In the following part, we mainly discuss the detailed features of the three categories of flaws, together with a few examples.
RTLLM~\cite{lu2024rtllm} flawed cases are not related to diagrams. \emph{cid002} and \emph{cid003} tasks in the CVDP~\cite{pinckney2025comprehensive} benchmark contain new flawed cases related to the syntax.\looseness=-1

\textbf{Re-evaluation on Updated Benchmarks.}
We re-evaluate the performance of advanced LLMs (e.g., GPT-4o~\cite{achiam2023gpt} and Claude-3.7) on the updated benchmark. 
Table~\ref{tab:Test-Benchmark} presents the {updated performance}. 
We can observe that Claude-3.7's performance shows minor variance, while GPT-4o~\cite{achiam2023gpt} presents an obvious improvement. This observation infers that the performance of GPT-4o is slightly underestimated. With the updated benchmark, we are able to conduct a fairer and more realistic evaluation of different LLMs.\looseness=-1

\begin{figure}[!b]
    \vspace{-.1in}
    \centering
    \includegraphics[width=.95\linewidth]{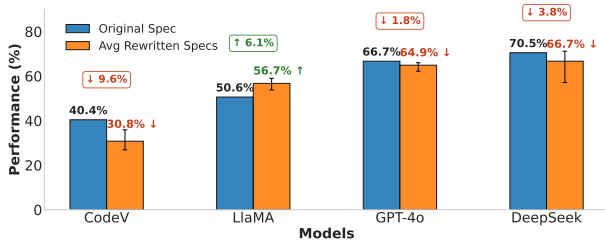}
    \vspace{-.1in}
    \caption{Performance evaluation based on the rewritten design descriptions (Rewritten Specs). CodeV, fine-tuned on the RTL generation tasks, shows more serious performance degradation, indicating more potential overfitting issues.}
    \label{fig:overfitting-detection}
\end{figure}

\subsection{Overfitting Detection}

In this section, we discuss the overfitting analysis results based on the performance after design description rewriting. For each original description, we generate four rewritten descriptions for evaluation. Figure~\ref{fig:overfitting-detection} illustrates the overall performance comparison results. From the figure, we can see that most of the models show lower performance than original description. Among the models, CodeV~\cite{zhao2024codev} shows a significantly lower performance (9.6\% decrease). The most advanced commercial models, GPT-4o~\cite{achiam2023gpt} and DeepSeek~\cite{liu2024deepseek}, both show slight degradation on performance. Interestingly, the open-sourced model LlaMA~\cite{touvron2023llama} demonstrates an obvious performance improvement (up to 6.1\%). Additionally, we collect the best and worst performance of LLMs on the four \colorbox{overfittingorange!14}{rewritten descriptions}. DeepSeek~\cite{liu2024deepseek} and CodeV~\cite{zhao2024codev}, with more degradation on performance, also show larger performance variance than LlaMA~\cite{touvron2023llama} and GPT-4o~\cite{achiam2023gpt}.

\begin{figure}[b!]
    \vspace{-.2in}
    \centering
    \includegraphics[width=0.95\linewidth]{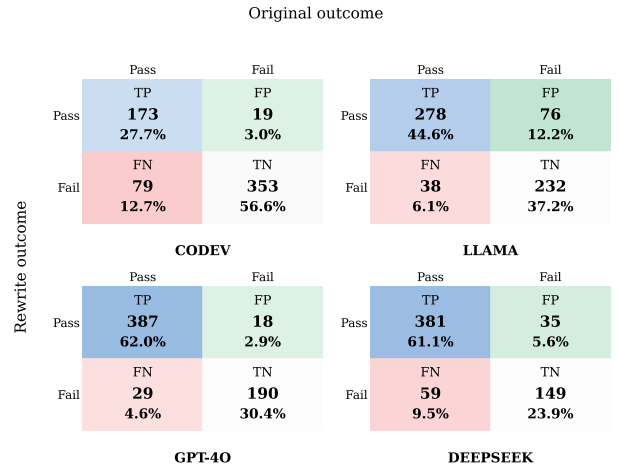}
    \vspace{-.1in}
    \caption{Overfitting analysis matrix. The most advanced open-sourced model, CodeV~\cite{zhao2024codev}, exhibits the most severe overfitting (FN), while the open-sourced foundation model LLaMA demonstrates the greatest improvement (FP).}
    \label{fig:overfitting-matrix}
\end{figure}

Figure~\ref{fig:overfitting-matrix} illustrates the overfitting analysis matrix for each model. 
\colorbox{softblue!14}{TP} items means the evaluation results on both original and updated descriptions are passed. The summation of \colorbox{softblue!14}{TP} and \colorbox{softgreen!14}{FP} reveals the overall performance on the updated descriptions.
We consider \colorbox{softgreen!14}{FP} a strong indicator of LLM generalizability, and \colorbox{softblue!14}{TP} a good indicator of LLM robustness.
The figure further demonstrates that CodeV~\cite{zhao2024codev} suffers the most serious overfitting (\colorbox{softRed!14}{FN}) problem. Here, (\colorbox{softRed!14}{FN}) means that LLMs pass on the original description while failing the updated descriptions. We consider this item a strong indicator of overfitting. On the other hand, CodeV~\cite{zhao2024codev}s LlaMA~\cite{touvron2023llama} demonstrates a significantly higher number for \colorbox{softgreen!14}{FP} items. Here, \colorbox{softgreen!14}{FP} means that the LLMs fail on the original description while passing the updated descriptions. 
Among the fundation models, DeepSeek~\cite{liu2024deepseek} shows higher overfitting (\colorbox{softRed!14}{FN}) problems over GPT-4o~\cite{achiam2023gpt} and LlaMA~\cite{touvron2023llama}.

\section{Conclusion}\label{sec:concl}

This work introduces RTL-BenchMT, an agentic framework for dynamically maintaining RTL generation benchmarks.
RTL-BenchMT targets two core applications: (1) \emph{automatically identifying and revising flawed cases}, and (2) \emph{automatically detecting and updating overfitting cases}.
With the assistance of RTL-BenchMT, we conduct an in-depth analysis of ambiguity and overfitting in existing benchmarks and show that a non-trivial fraction of LLM failures originates from benchmark flaws.
Finally, we contribute a refined benchmark suite, together with the accompanying agentic maintenance framework, which we will open-source to the community.

\section{Acknowledgement}

This work is supported by the Hong Kong Research Grants Council (RGC) CRF-YCRG C6003-24Y, GRF 16216825. It was partially conducted by ACCESS – AI Chip Center for Emerging Smart Systems, supported by the InnoHK initiative of the Innovation and Technology Commission of the Hong Kong Special Administrative Region Government.


\bibliographystyle{ACM-Reference-Format}
\bibliography{ref}

\end{document}